\documentclass[lettersize,journal]{IEEEtran}

\usepackage{amsmath,amsfonts}
\usepackage{algpseudocode,algorithm}
\usepackage{array}
\usepackage{textcomp}
\usepackage{stfloats}
\usepackage{url}
\usepackage{verbatim}
\usepackage{graphicx}
\usepackage{cite}

\usepackage{colortbl}
\usepackage{xcolor}
\usepackage{tabularx}
\usepackage{siunitx}
\usepackage{balance}
\ifCLASSOPTIONcompsoc
\usepackage[caption=false, font=normalsize, labelfont=sf, textfont=sf]{subfig}
\else
\usepackage[caption=false, font=footnotesize]{subfig}
\fi
\graphicspath{{./fig/}}

\usepackage{blindtext}
\usepackage{hyperref}

\begin{document}

\title{BiC-MPPI: Goal-Pursuing, Sampling-Based Bidirectional Rollout Clustering Path Integral for Trajectory Optimization}

\author{Minchan Jung and Kwang-Ki~K.~Kim$^*$
\thanks{The authors are with the ECE dept. at Inha University, Republic of Korea. This research was supported by the BK21 Four Program funded by the Ministry of Education (MOE, Korea) and National Research Foundation of Korea (NRF). ${}^{*}$Corresponding author ({\tt kwangki.kim@inha.ac.kr}).}
}

\maketitle

\begin{abstract}
This paper introduces the Bidirectional Clustered MPPI (BiC-MPPI) algorithm, a novel trajectory optimization method aimed at enhancing goal-directed guidance within the Model Predictive Path Integral (MPPI) framework. BiC-MPPI incorporates bidirectional dynamics approximations and a new guide cost mechanism, improving both trajectory planning and goal-reaching performance. By leveraging forward and backward rollouts, the bidirectional approach ensures effective trajectory connections between initial and terminal states, while the guide cost helps discover dynamically feasible paths. Experimental results demonstrate that BiC-MPPI outperforms existing MPPI variants in both 2D and 3D environments, achieving higher success rates and competitive computation times across 900 simulations on a modified BARN dataset for autonomous navigation.
\\[1mm] 
\indent{\rm GitHub:} \href{https://github.com/i-ASL/BiC-MPPI}{\tt https://github.com/i-ASL/BiC-MPPI}
\vspace{-2mm}
\end{abstract}



\section{Introduction}
\label{sec:intro}
Optimization-based trajectory generation methods have gained significant attention due to advancements in computational performance and optimization algorithms. In the past, trajectories were primarily generated using heuristic approaches such as gird-based motion planners (e.g., Breadth-First Search (BFS), Depth-First Search (DFS), Dijkstra, A$^*$ and D$^*$ algorithms) and sampling-based motion planners (e.g., Probabilistic Roadmaps (PRM, PRM$^*$), Rapidly-exploring Random Trees (RRT, RRT$^*$), and RRT variants). However, optimization-based trajectory generation techniques with consideration of kinematic or dynamic constraints are now being applied in real time across various industries~\cite{choset2005principles,lavalle2006planning,chai2020overview}.

Trajectory optimization not only generates optimal trajectories but also identifies optimal sequences of actions or inputs using a system model. These approaches can be broadly classified into gradient-based and sampling-based methods. Differential Dynamic Programming (DDP) addresses trajectory optimization through iterative forward and backward sweeps until a specified tolerance is achieved, utilizing gradients and Hessians around a nominal trajectory. With the rapid advancement of GPUs and parallel computation, sampling-based methods, such as Model Predictive Path Integral (MPPI), have gained significant attention. Recently, various techniques have been integrated into the MPPI framework~\cite{kazim2024recent}. However, sampling-based methods still face challenges in goal finding. While terminal costs can be applied, they are often limited to a specific target point.

Bidirectional path planning, known for its advantages in goal finding, typically involves solving two-point boundary value problems (BVPs)~\cite{jordan2013optimal}. However, deriving a closed-form solution is often infeasible, leading to the widespread use of numerical methods~\cite{osborne1969shooting,keller2018numerical}. Even without directly solving the BVP, forward rollouts can leverage heuristic information from backward rollouts~\cite{nayak2022bidirectional}. Some methods combine MPPI with other optimal control strategies, such as Interior Point Differential Dynamic Programming (IPDDP)~\cite{kim2024mppi} or ancillary controllers~\cite{trevisan2024biased}. These approaches require the integration of additional optimal controllers beyond the standard MPPI framework.

This paper proposes the Bidirectional Clustered MPPI (BiC-MPPI), an enhanced trajectory optimization method that improves goal-directed guidance within the MPPI framework. The main contributions of this work are as follows:
(i) \textit{Bidirectional Approach}: Different from existing MPPI variants that operate solely in the forward direction, our method introduces a novel bidirectional path integral approach, which integrates backward dynamics approximations for improved trajectory planning;
(ii) \textit{Guide Cost}: We extend the MPPI framework by incorporating a guide cost mechanism that goes beyond traditional cost functions, which typically rely on terminal costs or target-combined stage costs. This mechanism aids in discovering goal-reaching trajectories by considering dynamically feasible reference trajectories; and
(iii) \textit{Successful Goal Navigation}: Our approach demonstrates superior goal-oriented navigation performance, successfully navigating complex environments and outperforming existing MPPI algorithms in hundreds of test scenarios.



\section{Background}
\label{sec:bg}

\subsection{Trajectory Optimization}
\label{sec:bg:ocp}
For state-space trajectory optimization, we consider the state $x_t \in \mathbb{R}^n$ and the input $u_t \in \mathbb{R}^m$ with the stage cost $\psi: \mathbb{R}^{n} \times \mathbb{R}^{m} \rightarrow \mathbb{R}$ and the terminal state cost $\phi :\mathbb{R}^{n} \rightarrow \mathbb{R}$.
The associated optimal control problem (OCP) of trajectory optimization can be represented as the following:
\begin{equation}\label{eq:bg:ocp:1}
	\begin{aligned}
		\min_{u_{0:T-1}} \quad \ & \phi(x_T) + \sum_{t=0}^{T-1} \psi(x_t,u_t) \\
		\text{subject to} \quad 
		& x_{t+1} = f(x_t,u_t),\,x_0 = x_{\rm init}\\
		& g(x_t) \leq 0,\, p(x_t) \notin \mathcal{O} , \, h(u_t) \leq 0
	\end{aligned}
\end{equation}
where \( x_{\text{init}} \) represents the initial condition, \( f: \mathbb{R}^n \times \mathbb{R}^m \rightarrow \mathbb{R}^n \) denotes the control system dynamics, and the mappings \( g : \mathbb{R}^{n} \rightarrow \mathbb{R}^{\ell_x} \) and \( h : \mathbb{R}^{m} \rightarrow \mathbb{R}^{\ell_u} \) represent the state and input constraints, respectively. In addition to these functional state constraints, we also consider the collision avoidance constraint \( p(x_t) \notin \mathcal{O} \), where \( p : \mathbb{R}^{n} \rightarrow \mathbb{R}^{d} \) denotes the position of the state \( x_t \), and \( \mathcal{O} \) represents the obstacle region in the environment.

\subsection{Model Predictive Path Integral}
\label{sec:bg:mppi}
MPPI, a sampling-based trajectory optimization method, generates random trajectories (samples) with noise-injected inputs \( v_t = u_t + \varepsilon_t \), sequentially from the initial state \( x_{\rm init} \). The sequence of random noise \( \varepsilon_{0:T-1} = (\varepsilon_0, \cdots, \varepsilon_{T-1}) \) follows a normal distribution \( \mathcal{N}(0, \Sigma_u) \), with zero mean and covariance matrix \( \Sigma_u \in \mathbb{R}^{m \times m} \). We consider the input constraint set \( \mathbb{U} = \{ u_{0:T-1} \in \mathbb{R}^{mT} : h(u_t) \leq 0 \text{ for } t = 0, \ldots, T-1 \} \).

In sampling-based methods, a large number of trajectories are generated, and a weighted average of the sampled rollout trajectories is used to refine the final path or control input. This weighted averaging ensures that the resulting inputs not only meet specific state conditions but also effectively achieve the navigation goal, all while adhering to physical limits. To prevent the generation of inputs that violate constraints, a projection operator \( \mathcal{P}_{\mathbb{U}} : \mathbb{R}^{mT} \rightarrow \mathbb{U} \) is employed, defined as:
\begin{equation}
\mathcal{P}_{\mathbb{U}}(U) = \arg \min_{U' \in \mathbb{U}} \| U' - U \|_2 \,.
\end{equation}
Here, \( \mathbb{U} \) represents the input constraint set, which is assumed to be non-empty, closed, and convex. This assumption ensures that the projection \( \mathcal{P}_{\mathbb{U}}(U) \) always exists and is unique for any \( U \in \mathbb{R}^{mT} \). In certain cases, such as box-type constraints or second-order cone constraints, closed-form solutions for the projection are available~\cite{bauschke1996projection,kim2024mppi}. Since the input constraints are decoupled across time steps, the projection can be applied element-wise for each input \( u \in \mathbb{R}^{m} \), where the set \( \mathcal{U} = \{ u \in \mathbb{R}^{m} : h(u) \leq 0 \} \) is also closed and convex, thereby ensuring feasible inputs at each time step.


MPPI considers  soft constraints by incorporating hard constraints as scalar values within the cost function. It is important to note that the dynamic constraints and initial condition are naturally satisfied through forward rollouts using the dynamics \( f \) from \( x_{\rm init} \). The cost function is defined as:
\begin{equation}\label{eq:mppi:cost}
	\min_{U} \ \  J(U) = \phi (x_T) + \sum^{T-1}_{t=1} \left\{ \psi(x_t,u_t) + \mathcal{I}_{\mathcal{X}}(x_t) \right\}
\end{equation}
where the collision checker \( \mathcal{I}_{\mathcal{X}} \) is defined by the following indicator function:
\begin{equation}
    \mathcal{I}_{\mathcal{X}}(x_t) =
    \begin{cases}
        \infty, & \text{if } p(x_t) \in \mathcal{O} \, \lor \, g(x_t) > 0  \\
        0, & \text{otherwise}
    \end{cases}
\end{equation}
Here, the set \( \mathcal{X} = \{ x \in \mathbb{R}^{n} : g(x) \leq 0, \, p(x_t) \notin \mathcal{O} \} \) denotes the combined state constraint set, which accounts for both the state constraints \( g(x_t) \leq 0 \) and the collision avoidance condition \( p(x_t) \notin \mathcal{O} \).


Since a sampled input may violate the input constraints, each sampled input sequence \( U^k=(u_0^k,\dots,u_{T-1}^k)\), where each \( u_t^{k} \sim \mathcal{N}(\bar{u}_t, \Sigma_u) \), is first projected onto the closed convex set \( \mathcal{U} \) to discard infeasible sampled trajectories. This projection is applied as \( U^{k} \leftarrow \mathcal{P}_{\mathcal{U}}(U^{k}) \). After generating trajectories with these projected inputs and computing the corresponding costs, the relative weight of each trajectory is calculated. A weighted average of the sampled inputs, forming a convex combination, is then used to determine the optimal control input sequence \( U^* = (u_0^*, \dots, u_{T-1}^*) \):\\[-2mm]
\begin{equation}\label{eq:weightsuminput}
w^k = \exp(-\gamma_u J^k), \ \, \bar{w} = \sum^{N_{\rm s}}_{k=1} w^k, \ \, U^* = \sum^{N_{\rm s}}_{k=1} \frac{w^k}{\bar{w}} U^k
\end{equation}
where the right-superscript \( k \) is the sampling index and \( N_{\rm s} \) is the number of samples. Since each \( u^{k}_{t} \in \mathcal{U} \) for all \( k \) and \( t \), and \( \mathcal{U} \) is convex, the resulting inputs \( u_t^* \in \mathcal{U} \) are feasible for all \( t = 0, \dots, T-1 \), ensuring that \( U^* \in \mathbb{U} \).

\subsection{Rollout Clustering MPPI}
\label{sec:bg:clusterMppi}
Although numerous sampled trajectories are generated, the original naive MPPI typically yields a single path. This approach has limitations in efficiently handling samples when multiple local minima are present. For instance, if a robot encounters an obstacle, both turning left and turning right might be optimal trajectories with identical or similar costs, representing two different local minima. However, since the weighted average calculation \eqref{eq:weightsuminput} considers all samples simultaneously, the resulting trajectory may not necessarily lie within a feasible region. This can lead to situations where \( X(U^*) = (x_0, f(x_0, u_0^*), f(f(x_0, u_0^*), u_1^*), \cdots, f(..., u_{T-1}^*)) \notin \mathcal{X}^{T+1} \), indicating that the trajectory might fall outside the feasible state constraint set \( \mathcal{X} \), even though each individual sampled trajectory is feasible.

To avoid generating infeasible trajectories while leveraging the benefits of all sampled trajectories, rollout clustering MPPI is proposed in~\cite{patrick2024path}. The key idea of trajectory clustering can be summarized as follows: After generating and evaluating sample trajectories similarly to the naive MPPI presented in Section~\ref{sec:bg:mppi}, the DBSCAN (Density-Based Spatial Clustering of Applications with Noise)~\cite{ester1996density} algorithm is applied to cluster the rollout trajectory samples based on the deviations of input sequences and the associated costs. DBSCAN has two main parameters: the minimum number of points \( p \) required to define core points, and the maximum distance \( \epsilon_{\max} \) used to determine the neighborhood within a specified boundary.

For \( N_{\rm s} \) trajectory samples, consider the power set \( 2^{\mathcal{N}_{s}} \), which represents all subsets of \( \mathcal{N}_{s} \), including the empty set and the set itself. Here, \( \mathcal{N}_{s} = \{1, 2, \ldots, N_{\rm s}\} \) is the index set for sampling. The output of the clustering algorithm is a set of clusters \( \{ C_1, \dots, C_I \} \), where the clusters satisfy the disjoint-set property \( C_{i} \cap C_{j} = \emptyset \) for all \( i \neq j \), and the complete-set property \( \cup_{i=1}^{I} C_{i} = \mathcal{N}_{s} \). Each cluster \( C_i \in 2^{\mathcal{N}_{s}} \) contains the indices of the \( i \)-th group of clustered trajectories.
If no valid clusters are found due to incorrect parameter settings, a fallback strategy is applied by setting \( C_I = \mathcal{N}_{s} \) with \( I = 1 \), including all samples in a single cluster. For each cluster set, the optimal control \( U^{C_i} \) for the \( i \)-th cluster is computed as in~\eqref{eq:weightsuminput}, where the weighted sum is calculated over \( k \in C_i \) for each cluster \( i \).
Once the clusters are formed, the overall optimal control input is determined as:
\begin{equation}\label{eq:clusteroptimal}
U^{C^*} \gets\underset{U^{C_1},\cdots,U^{C_{\! I}}}{\arg\min}\,J(U^{C_i}) \,.
\end{equation}

\subsection{Bidirectional Rollout Kinematics}
\label{sec:bg:rolloutKinec}
To rollout with a sequence of control inputs over a finite horizon with sampling-time intervals, we need a simulation model of the control system corresponding to the kinematics or dynamics. 

\paragraph{Forward Rollout Kinematics}
For forward rollouts using a sampled control input sequence \( U^k \), any standard numerical method for solving ordinary differential equations (ODEs), such as \( \dot{x}(s) = F(x(s), u(s)) \) with the initial condition \( x(0) = x_{\rm init} \), can be employed~\cite{butcher2016numerical}. In this work, we use the well-known fourth-order Runge-Kutta (RK4) method, defined as follows:
\begin{equation}\label{eq:bg:rolloutKinec:RK4fore}
\begin{split}
\kappa_1 &= \delta{t}  F(x_t, u_t), \
\kappa_2 = \delta{t} F(x_t \!+\! {\kappa_1}/{2}, u_{\frac{t+1}{2}}), \\
\kappa_3 &= \delta{t} F(x_t \!+\!  {\kappa_2}/2, u_{\frac{t+1}{2}}), \
\kappa_4 = \delta{t} F(x_t \!+\!  \kappa_3, u_{t+1}) \\
x_{t+1} &= x_t + (\kappa_1 + 2\kappa_2 + 2\kappa_3 + \kappa_4) / 6 \\
\end{split}
\end{equation}
where \( x_{t} = x(t \delta t) \), \( u_{t} = u(t \delta t) \), and \( u_{\frac{t+1}{2}} = (u_{t} + u_{t+1}) / 2 \). This formulation slightly deviates from the control system model given in~\eqref{eq:bg:ocp:1}, but sampling-based MPPI methods do not require strict adherence to the Markovian nature of the system. Instead, they focus on computing the total cost of a rollout trajectory, given a sampled input sequence over the horizon. 

\paragraph{Backward Rollout Kinematics}
In our method of bidirectional rollout-based trajectory optimization, the backward dynamics are also required by solving the ordinary differential equation \( \dot{x}(s) = F(x(s), u(s)) \) with a final condition \( x(T) = x_{\rm goal} \). Similar to the forward RK4 method, the following backward Runge-Kutta (RK4) method is used for backward rollouts with a sampled sequence of control inputs \( U^k \):
\begin{equation}\label{eq:bg:rolloutKinec:RK4back}
\begin{split}
\kappa_1 &= \delta{t}  F(x_t, u_t), \
\kappa_2 = \delta{t} F(x_t \!-\! {\kappa_1}/{2}, u_{\frac{t-1}{2}}), \\
\kappa_3 &= \delta{t} F(x_t \!-\!  {\kappa_2}/2, u_{\frac{t-1}{2}}), \
\kappa_4 = \delta{t} F(x_t \!-\!  \kappa_3, u_{t-1}) \\
x_{t+1} &= x_t - (\kappa_1 + 2\kappa_2 + 2\kappa_3 + \kappa_4) / 6 \\
\end{split}
\end{equation}
where \( u_{\frac{t-1}{2}} = (u_{t} + u_{t-1}) / 2 \). The computation starts with the final condition \( x(T) = x_{\rm goal} \) and proceeds backward through the time steps, \( t = T, T-1, \ldots, 0 \).

\section{BiC-MPPI Algorithm}
\label{sec:BiC-MPPI}

%

This section introduces an extension of classical MPPI, the Bidirectional Clustered MPPI (BiC-MPPI) algorithm, which addresses the limitations of existing MPPI variants and enhances goal-directed trajectory optimization. 
%
%
The BiC-MPPI algorithm, whose pseudocode is provided in Algorithm~\ref{alg:bicmppi}, operates in three main stages:
\begin{enumerate}
\item
\emph{Bidirectional Clustered Trajectory Generation}: In this step, backward and forward trajectory branches are generated using the Clustered MPPI approach. However, instead of directly finding the optimal control input using~\eqref{eq:clusteroptimal}, multiple candidate trajectories are generated. These trajectories provide a broad exploration of the solution space, allowing for more robust trajectory options.
\item
\emph{Connection of Trajectory Branches}: Once the trajectories are generated, the next step is to select the best connection point between the forward and backward branches. This is done by calculating the distance between the state of the forward branch and the backward branch. The control inputs and states are then concatenated at the joint to form a single, continuous trajectory.
\item
\emph{Guide MPPI}: In the final step, a modified version of MPPI, called Guide MPPI, is applied. Here, a guide cost is introduced to encourage the trajectory to follow a reference trajectory generated from the concatenated forward-backward connection. This refines the trajectory and ensures that it smoothly pursues the desired path toward the goal, improving overall performance and reducing deviations from the optimal path.
\end{enumerate}
By combining these three steps, the BiC-MPPI algorithm enhances the standard MPPI framework, particularly in scenarios with complex dynamics and multiple local minima.

\begin{algorithm}[t]\small
	\caption{${\tt BiC\text{-}MPPI}$}\label{alg:bicmppi}
	\begin{algorithmic}[1]\small
		\State \textbf{Input:} initial state $x_{\rm init}$, goal state $x_{\rm goal}$
		\State \textbf{Output:} locally optimal control $U^*$
		\State \textbf{Initialize:} initial controls $U_{\rm f0}$, $U_{\rm b0}$
		\While{task is not finished}
            \State \textbf{Step 1. Clustered MPPI}
			\State $U_{\rm f}=\{U_{\rm f}^{i}\}_{i=1}^{{\rm I}_{\rm f}} \leftarrow {\tt ForwardCluster}(U_{\rm f0})$
            \Comment{Alg.~\ref{alg:forwardclusteredmppi}}
			\State $U_{\rm b}=\{U_{\rm b}^{i}\}_{i=1}^{\rm I_{\rm b}} \leftarrow {\tt BackwardCluster}(U_{\rm b0})$
            \Comment{Alg.~\ref{alg:backwardclusteredmppi}}
            \State \textbf{Step 2. Trajectory Association}
            \State $\{\check{U}^{(i)},\check{X}^{(i)}\}_{i=1}^{S} \leftarrow {\tt Connect}(U_{\rm b}, U_{\rm f})$
            \Comment{Sec.~\ref{sec:BiC-MPPI:connect}}
            \State \textbf{Step 3. Guide MPPI}
		  \For{$i = 1,\dots,S$}  \texttt{// Parallel}
                \State $U_{\rm g}^{(i)*} \leftarrow {\tt GuideMPPI}(\check{U}^{(i)},\check{X}^{(i)})$
                \Comment{Alg.~\ref{alg:guidedmppi}}
            \EndFor
            \State $U^* = \underset{U_{\rm g}^{(i)*}=\{U_{\rm g}^{(1)*},\dots,U_{\rm g}^{(S)*}\}}{\text{argmin}}\,J_{\rm f}(U_{\rm g}^{(i)*})$
            \Comment{(\ref{eq:clustermppiforwardcost})}
		\EndWhile
	\end{algorithmic}
\end{algorithm}

\subsection{Forward \& Backward Clustered MPPI}
\label{sec:BiC-MPPI:fbc}
The first step of the proposed BiC-MPPI is to generate bidirectional clustered trajectories using the forward and backward, i.e., bidirectional, clustered MPPI methods. Our bidirectional clustered trajectory generation works similar to the rollout clustering MPPI proposed in~\cite{patrick2024path}, but a main difference is that the results of all clustered trajectories are stored, instead of selecting only one optimal branch among clusters. The following OCP is considered for the forward clustered MPPI strategy:
\begin{equation}\label{eq:clustermppiforwardcost}
	\begin{aligned}
		\min_{U \in \mathbb{U}}  \ \  & J_{\rm f}(U) = \phi_{\rm f} (x_{T_{\rm f}}) + \sum^{T_{\rm f}-1}_{t=0} \left\{ \psi_{\rm f} (x_t, u_t) + \mathcal{I}_{\mathcal{X}_{\rm f}}(x_t) \right\} \\
		\text{s.t.} \quad 
		& x_{t+1} = f_{\rm f}(x_t,u_t),\,x_{0} = x_{\rm init}\\
	\end{aligned}
\end{equation}
where an initial condition $x_{\rm init}$ is given and the discrete-time model $f_{\rm f}(\cdot, \cdot)$ is obtained from the forward RK4 method presented in~\eqref{eq:bg:rolloutKinec:RK4fore}.
The cost functions about terminal and stage state ($\phi_{\rm f} : \mathbb{R}^{n}\rightarrow \mathbb{R}$ and $\psi_{\rm f} : \mathbb{R}^{n} \times \mathbb{R}^{m} \rightarrow \mathbb{R}$), and the state-constraint set $\mathcal{X}_{\rm f} \subset \mathbb{R}^{n}$ are defined same as the classical MPPI described in Section~\ref{sec:bg:mppi}. 

Same as the forward clustered MPPI, the resulting clusters and their information are stored for the use of connecting the forward and backward rollout clusters of trajectories in the later steps of the BiC-MPPI algorithm. 
We consider the following OCP for the backward clustered MPPI strategy:
\begin{equation}\label{eq:clustermppibackwardcost}
	\begin{aligned}
		\min_{U}  \ \  & J_{\rm b}(U) = \phi_{\rm b} (x_0) + \sum^{1}_{t=T_{\rm b}} \left\{ \psi_{\rm b} (x_t, u_t) + \mathcal{I}_{\mathcal{X}_{\rm b}}(x_t) \right\} \\
		\text{s.t.} \quad 
		& x_{t-1} = f_{\rm b}(x_t,u_t),\,x_{T_{\rm b}} = x_{\rm goal}\\
	\end{aligned}
\end{equation}
where $x_{\rm goal} \in \mathcal{X}_{\rm b}$ is a given targeted goal state and the discrete-time model $f_{\rm b}(\cdot, \cdot)$ is obtained from the backward RK4 method presented in~\eqref{eq:bg:rolloutKinec:RK4back}.
The cost functions about terminal and stage state ($\phi_{\rm b} : \mathbb{R}^{n}\rightarrow \mathbb{R}$ and $\psi_{\rm b} : \mathbb{R}^{n} \times \mathbb{R}^{m} \rightarrow \mathbb{R}$), and the state-constraint set $\mathcal{X}_{\rm b} \subset \mathbb{R}^{n}$ are defined similarly as the OCP for the forward clustered MPPI. 

%

\begin{algorithm}[t]\small
	\caption{${\tt Forward\ Clustered\ MPPI}$}\label{alg:forwardclusteredmppi}
	\begin{algorithmic}[1]\small
		\State \textbf{Input:} initial controls $U_{\rm 0}$
		\State \textbf{Output:} forward control inputs $U_{\rm f}$
		\State \textbf{Define:} $U=\{U^{k}\}_{k=1}^{N_{\rm f}}, J=\{J^{k}\}_{k=1}^{N_{\rm f}}, \varepsilon=\{\varepsilon^k\}_{k=1}^{N_{\rm f}}$
		\For{$k = 1,\dots,N_{\rm f}$} \texttt{// Parallel}
            \State $\varepsilon^k = \varepsilon^k_{0:T_{\rm f}-1} \ \text{where} \ \varepsilon^k_i\sim\mathcal{N}(0,\Sigma_u), \ i \in \mathbb{Z}_{0:T_{\rm f}-1}$
            \State $U^k \leftarrow \mathcal{P}_{\mathcal{U}}(U_0 + \varepsilon^k)$ , \ $J^k = J_{\rm f}(U^k)$
            \Comment{(\ref{eq:clustermppiforwardcost})}
        \EndFor
        \State $\mathcal{C} = \{ {C}_{1}, \dots, {C}_{\rm I_f} \} \leftarrow {\tt DBSCAN}(\varepsilon, J)$
        \If{cluster is not found}
            \State $C_{\rm I_f} = \mathcal{N}_{\rm f} \ \text{with}\ {\rm I_f} = 1$
        \EndIf
        \State \textbf{Define:} $U_{\rm f}=\{U_{\rm f}^{i}\}_{i=1}^{\rm I_f}, w=\{w^k\}_{k=1}^{N_{\rm f}}$
        \For{$\text{each} \ C \ \text{in} \ \mathcal{C}$}
            \State $\bar{J} = \underset{i \in C}{{\min}} (J^i)$
            \For{$\text{each} \ s \ \text{in} \ C$}
                \State $J^{s} = J^{s} - \bar{J}$, \ $w^{s} = \exp({-\gamma_u}J^{s})$
            \EndFor
            \State $\bar{w} = \sum_{s \in C} w^{s}$, \ $U_{\rm f}^i = \sum_{s \in C} (w^{s}/\bar{w})U^s$
        \EndFor
    \end{algorithmic}
\end{algorithm}

\begin{algorithm}[t]\small
        \caption{${\tt Backward\ Clustered\ MPPI}$}\label{alg:backwardclusteredmppi}
        \begin{algorithmic}[1]\small
		\State \textbf{Input:} initial controls $U_{\rm 0}$
		\State \textbf{Output:} backward control inputs $U_{\rm b}$
		\State \textbf{Define:} $U=\{U^{k}\}_{k=1}^{N_{\rm b}}, J=\{J^{k}\}_{k=1}^{N_{\rm b}}, \varepsilon=\{\varepsilon^k\}_{k=1}^{N_{\rm b}}$
		\For{$k = 1,\dots,N_{\rm b}$} \texttt{// Parallel}
            \State $\varepsilon^k = \varepsilon^k_{0:T_{\rm b}-1} \ \text{where} \ \varepsilon^k_i\sim\mathcal{N}(0,\Sigma_u), \ i \in \mathbb{Z}_{0:T_{\rm b}-1}$
            \State $U^k \leftarrow \mathcal{P}_{\mathcal{U}}(U_0 + \varepsilon^k)$, \ $J^k = J_{\rm b}(U^k)$
            \Comment{(\ref{eq:clustermppibackwardcost})}
        \EndFor
        \State $\mathcal{C} = \{ {C}_{1}, \dots, {C}_{\rm I_b} \} \leftarrow {\tt DBSCAN}(\varepsilon, J)$
        \If{cluster is not found}
            \State $C_{\rm I_b} = \mathcal{N}_{\rm b} \ \text{with}\ {\rm I_b} = 1$
        \EndIf
        \State \textbf{Define:} $U_{\rm b}=\{U_{\rm b}^{i}\}_{i=1}^{\rm I_b}, w=\{w^k\}_{k=1}^{N_{\rm b}}$
        \For{$\text{each} \ C \ \text{in} \ \mathcal{C}$}
            \State $\bar{J} = \underset{i \in C}{{\min}} (J^i)$
            \For{$\text{each} \ s \ \text{in} \ C$}
                \State $J^{s} = J^{s} - \bar{J}$, \ $w^{s} = \exp({-\gamma_u}J^{s})$
            \EndFor
            \State $\bar{w} = \sum_{s \in C} w^{s}$, \ $U_{\rm b}^i = \sum_{s \in C} (w^{s}/\bar{w})U^s$
        \EndFor
    \end{algorithmic}
\end{algorithm}

\subsection{Trajectory Association for Connection}
\label{sec:BiC-MPPI:connect}
The previous step generates multiple trajectory candidates in both the forward and backward directions, starting from the initial condition \( x_{\rm init} \) and the terminal condition \( x_{\rm goal} \), respectively. Assuming that none of these trajectories directly connects the initial and terminal conditions, the next step involves associating or pairing opposite directional branches of trajectory candidates. This ensures a complete trajectory that bridges \( x_{\rm init} \) and \( x_{\rm goal} \).

In this trajectory association process, the nearest points between the forward and backward trajectory branches are identified based on a chosen distance metric. After determining the closest points, each trajectory is cut at its nearest point, and the trimmed forward and backward trajectories are connected at these points, without considering kinematic or state constraints at this stage. This straightforward connection acts as an intermediate step before further refining the overall trajectory in subsequent stages.

Let \( \mathcal{C}^{\rm f} = \{ {C}^{\rm f}_{1}, \dots, {C}^{\rm f}_{\rm I_f} \} \) and \( \mathcal{C}^{\rm b} = \{ {C}^{\rm b}_{1}, \dots, {C}^{\rm b}_{\rm I_b} \} \) represent the sets of forward and backward trajectory clusters, respectively, generated in the previous step. Each cluster corresponds to a set of sampled trajectories that share similar costs and states.
For the forward trajectories, the state trajectory corresponding to the \( i \)-th forward cluster is denoted by:
\(
X^{C^{\rm f}_{i}} = ({}^{\rm f\!}x^{i}_0, {}^{\rm f\!}x^{i}_1, \dots, {}^{\rm f\!}x^{i}_{T_{\rm f}})
\)
where \( {}^{\rm f\!}x^{i}_t \) represents the state at time \( t \) in the \( i \)-th forward trajectory cluster, and \( T_{\rm f} \) is the time horizon for the forward rollout.
Similarly, for the backward trajectories, the state trajectory corresponding to the \( j \)-th backward cluster is:
\(
X^{C^{\rm b}_{j}} = ({}^{\rm b\!}x^{j}_{0}, \ {}^{\rm b\!}x^{j}_1, \ \dots, \ {}^{\rm b\!}x^{j}_{T_{\rm b}})
\)
where \( {}^{\rm b\!}x^{j}_t \) represents the state at time \( t \) in the \( j \)-th backward trajectory cluster, and \( T_{\rm b} \) is the time horizon for the backward rollout. The backward trajectories begin at \( {}^{\rm b\!}x^{j}_{T_{\rm b}} \) ($= x_{\rm goal}$) and proceed to \( {}^{\rm b\!}x^{j}_{0} \) (the closest point to the initial state). 
The input trajectories, denoted as \( U^{C^{\rm f}_{i}} \) and \( U^{C^{\rm b}_{j}} \), are defined similarly for both forward and backward clusters, following the respective control inputs applied at each time step.

For each forward branch \( C^{\rm f}_{i} \), the goal is to find an associated backward branch by solving the following optimization problem:
\begin{equation}\label{eq:nearest}
\begin{split}
    ( j_i , \tau_i, \tau_{j_i} )  = &\,  \underset{(j,\tau, \tau' )}{\arg\min} \,  \| {}^{\rm f\!}x^{i}_{\tau} - {}^{\rm b\!}x^{j}_{\tau'} \| \\
    &\, \text{s.t.} \, j \in \mathbb{Z}_{\rm 1:I_b}, \, \tau \in \mathbb{Z}_{0:T_{\rm f}}, \, \tau' \in \mathbb{Z}_{0:T_{\rm b}}
\end{split}
\end{equation}
where \( \mathbb{Z}_{I_{\rm l}:I_{\rm r}} = \{I_{\rm l},I_{\rm l}+1,\dots,I_{\rm r}\} \) is a set of integers between the integers $I_{\rm l}$ and $I_{\rm r}$. The objective is to minimize the distance between the forward trajectory state \( {}^{\rm f\!}x^{i}_{\tau} \) and the backward trajectory state \( {}^{\rm b\!}x^{j}_{\tau'} \), using a chosen distance metric such as Euclidean, Manhattan, or LQR cost metrics~\cite{tedrake2009lqr} to identify the closest points on the forward and backward trajectories.

Once the optimization problem~\eqref{eq:nearest} is solved, the selected backward branch is denoted as \( \hat{X}^{C^{\rm b}_{j_i}} \), which is a trimmed version of the backward cluster \( {X}^{C^{\rm b}_{j_i}} \). This branch starts at the identified cut point \( {}^{\rm b\!}x^{j}_{\tau_{j_i}} \) and extends to the goal state \( {}^{\rm b\!}x^{j}_{T_{\rm b}} = x_{\rm goal} \). Similarly, \( \hat{X}^{C^{\rm f}_{i}} \) represents the forward branch connecting the initial state \( {}^{\rm f\!}x^{i}_{0} = x_{\rm init} \) to the cut point \( {}^{\rm f\!}x^{i}_{\tau_i} \). The corresponding trimmed input trajectories follow the same pattern as the state trajectories.


The result of this trajectory association is the concatenation of the state and control input sequences:
\begin{equation}\label{eq:bi-connect}
\check{U}^{(i)} = \hat{U}^{C^{\rm f}_{i}}_{0:\tau_i-1} \oplus \hat{U}^{C^{\rm b}_{j_i}}_{\tau_{j_i}:T_{\rm b}-1}, \
\check{X}^{(i)} = \hat{X}^{C^{\rm f}_{i}}_{0:\tau_i} \oplus \hat{X}^{C^{\rm b}_{j_i}}_{\tau_{j_i}+1:T_{\rm b}}
\end{equation}
where \( \oplus \) represents the concatenation of two matrices with compatible dimensions. \( Y_{a:b} \) denotes the sequence \( (y_a, y_{a+1}, \dots, y_b) \), which defines the portion of the trajectories included in the concatenation.

The time horizon for the concatenated trajectory \( {T}_i = \tau_i + (T_{\rm b} - \tau_{j_i}) \) may be shorter than the original forward horizon \( T_{\rm f} \). To handle this, we append a dummy input (typically a zero-input) at the end of \( \check{U}^{(i)} \), and the target state \( x_{\rm goal} \) is appended to the end of \( \check{X}^{(i)} \). This ensures that the trajectory maintains a consistent time horizon of at least \( T_{\rm f} \), preserving the stability and effectiveness of the optimization process.
The use of the concatenated sequences \( \check{U}^{(i)} \) and \( \check{X}^{(i)} \) will be elaborated upon in the next section, where they are integrated into the Guide MPPI for further optimization.

\subsection{Guide MPPI}
\label{sec:BiC-MPPI:guide}
In traditional MPPI, these terminal costs serve as distant targets, essentially guiding the system with minimal information along the trajectory. This makes it difficult for sampling-based methods to efficiently generate goal-reaching trajectories, particularly in environments with complex obstacles or dynamics. Furthermore, when the cost function is overly focused on the goal state at a specific time step, it can lead to instability and render the cost function unreliable.
Guide MPPI mitigates these issues by incorporating a reference trajectory generated from the concatenated forward and backward branches. The guide cost incentivizes the sampled trajectories to follow this reference, encouraging more accurate and smooth transitions from the initial state to the goal. This method provides intermediate guidance throughout the entire horizon, improving trajectory quality while maintaining the flexibility of sampling-based optimization.

For each $i$-th concatenated clustered trajectory, we define the following guided trajectory optimization:
\begin{equation}\label{eq:BiC-MPPI:guide:mppi}
	\begin{aligned}
		\min_{U \in \mathbb{U}}  \ \  & J_{\rm g}^{(i)} ( U ; \check{U}^{(i)}, \check{X}^{(i)}) + \frac{\|  x_{T_i} - x_{\rm goal}\|}{\epsilon} \\
		\text{s.t.} \quad 
		& x_{t+1} = f_{\rm f}(x_t,u_t), \, t=0,1,\ldots, T_i ; \,  x_{0} = x_{\rm init}
	\end{aligned}
\end{equation}
where \( J_{\rm g}^{(i)} ( U ; \check{U}^{(i)}, \check{X}^{(i)}) = J(U) + d((X,U),(\check{X}^{(i)},\check{U}^{(i)})) \) and \( \epsilon >0 \) is a small inverse-weight.
Here, \( J(U) \) is defined similarly to the classical MPPI cost~(\ref{eq:mppi:cost}), but with the time horizon \( T_i \), and it combines state constraints over the intersection \( \mathcal{X} = \mathcal{X}_{\rm f} \cap \mathcal{X}_{\rm b} \). The terminal, initial, and stage costs are all included, effectively summing the forward and backward costs, \( J_{\rm f}(\cdot) \) and \( J_{\rm b}(\cdot) \), as defined in~\eqref{eq:clustermppiforwardcost} and~\eqref{eq:clustermppibackwardcost}, respectively.
The term \( d((X,U),(\check{X}^{(i)},\check{U}^{(i)})) \) represents a penalty for deviations from the guided trajectory \( (\check{X}^{(i)}, \check{U}^{(i)}) \), encouraging the trajectories to stay close to this reference\footnote{This distance metric can be decoupled as \( d((X,U),(\check{X}^{(i)},\check{U}^{(i)})) = \lambda_x d(X,\check{X}^{(i)}) + \lambda_u d(U,\check{U}^{(i)}) \), where \( \lambda_x, \lambda_u \geq 0 \) are weights for following the reference state and input trajectories, respectively. Setting \( \lambda_u = 0 \) can highlight the state trajectory association without biasing the control input.}.
By utilizing information from the previous MPPI stage, this penalty helps ensure that the resulting trajectory is feasible and closely follows the desired path from \( x_{\rm init} \) to \( x_{\rm goal} \).

To derive the optimal control with the guide cost \( J_{\rm g}^{(i)} \), we substitute \( J \) with \( J_{\rm g}^{(i)} \) in~(\ref{eq:mppi:cost}) and~(\ref{eq:weightsuminput}), allowing us to calculate the optimal trajectory for each selection with the optimal input \( U^{(i)*}_{\rm g} \). After evaluating all selections, the overall optimal input \( U^* \) is determined using~(\ref{eq:clusteroptimal}) and the optimal control problem~\eqref{eq:clustermppiforwardcost}. Importantly, this additional computation is efficient because the comparison is made at the same time step between \( (X, U) \) and \( (\check{X}, \check{U}) \), avoiding the need for a double loop over the entire time horizon.

\begin{algorithm}[t]\small
	\caption{${\tt Guide\ MPPI}$}\label{alg:guidedmppi}
	\begin{algorithmic}[1]\small
		\State \textbf{Input:} warm-start input $\check{U}$, reference trajectories ($\check{U}, \check{X}$)
		\State \textbf{Output:} guided control input $U_{\rm g}$
  		\State \textbf{Define:} $U=\{U^{k}\}_{k=1}^{N_{\rm g}}, J=\{J^{k}\}_{k=1}^{N_{\rm g}}, w=\{w^k\}_{k=1}^{N_{\rm g}}$
		\For{$k = 1,\dots,N_{\rm g}$} \texttt{// Parallel}
            \State $\varepsilon^k = \varepsilon^k_{0:T_{\rm g}-1} \ \text{where} \ \varepsilon^k_i\sim\mathcal{N}(0,\Sigma_u), \ i \in \mathbb{Z}_{0:T_{\rm g}-1}$
            \State $U^k \leftarrow \mathcal{P}_{\mathcal{U}}(U_0 + \varepsilon^k)$, \ $J^k = J_{\rm g}(U^k; \check{U}, \check{X})$
            \Comment{(\ref{eq:BiC-MPPI:guide:mppi})}
        \EndFor
        \State $\bar{J} = \min (J)$
		\For{$k = 1,\dots,N_{\rm g}$}
            \State $J^{k} = J^{k} - \bar{J}$, \  $w^{k} = \exp({-\gamma_u}J^{k})$
        \EndFor 
        \State $\bar{w} = \sum^{N_{\rm g}}_{k=1} w^{k}$, \ $U_{\rm g} = \sum^{N_{\rm g}}_{k=1} (w^{k}/\bar{w})U^k$
	\end{algorithmic}
\end{algorithm}

\section{Simulation Experiment}
\label{sec:sim}

\subsection{Autonomous Ground Vehicle Path Planning}
To numerically evaluate the performance of our approach, we conducted simulations in which a differential wheeled mobile robot pursued a given target. The system model for the 2D wheeled mobile robot is described by the following continuous equations:
\begin{equation*}
	\begin{aligned}
		\dot{x} = v_t \cos(\theta_t), \quad
		\dot{y} = v_t \sin(\theta_t), \quad
		\dot{\theta} = w_t,
	\end{aligned}
\end{equation*}
where \( x \) and \( y \) represent the robot's position, and \( \theta \) denotes its heading angle. The control inputs are linear velocity \( v_t \) and angular velocity \( w_t \), forming the control vector \( u_t = [v_t, w_t]^\top \).
The following constraints are considered for the OCP:
\begin{equation*}
	\begin{aligned}
		0 \le v_t \le 1.0, \quad
		|w_t| \le \frac{\pi}{4}, \quad
		(x_t, y_t) \notin \mathcal{O},
	\end{aligned}
\end{equation*}
where the obstacle region \( \mathcal{O} \) is defined based on the environment configuration.

\begin{figure}[t]
    \centering
	\includegraphics[width=.925\linewidth, height=0.4\linewidth]{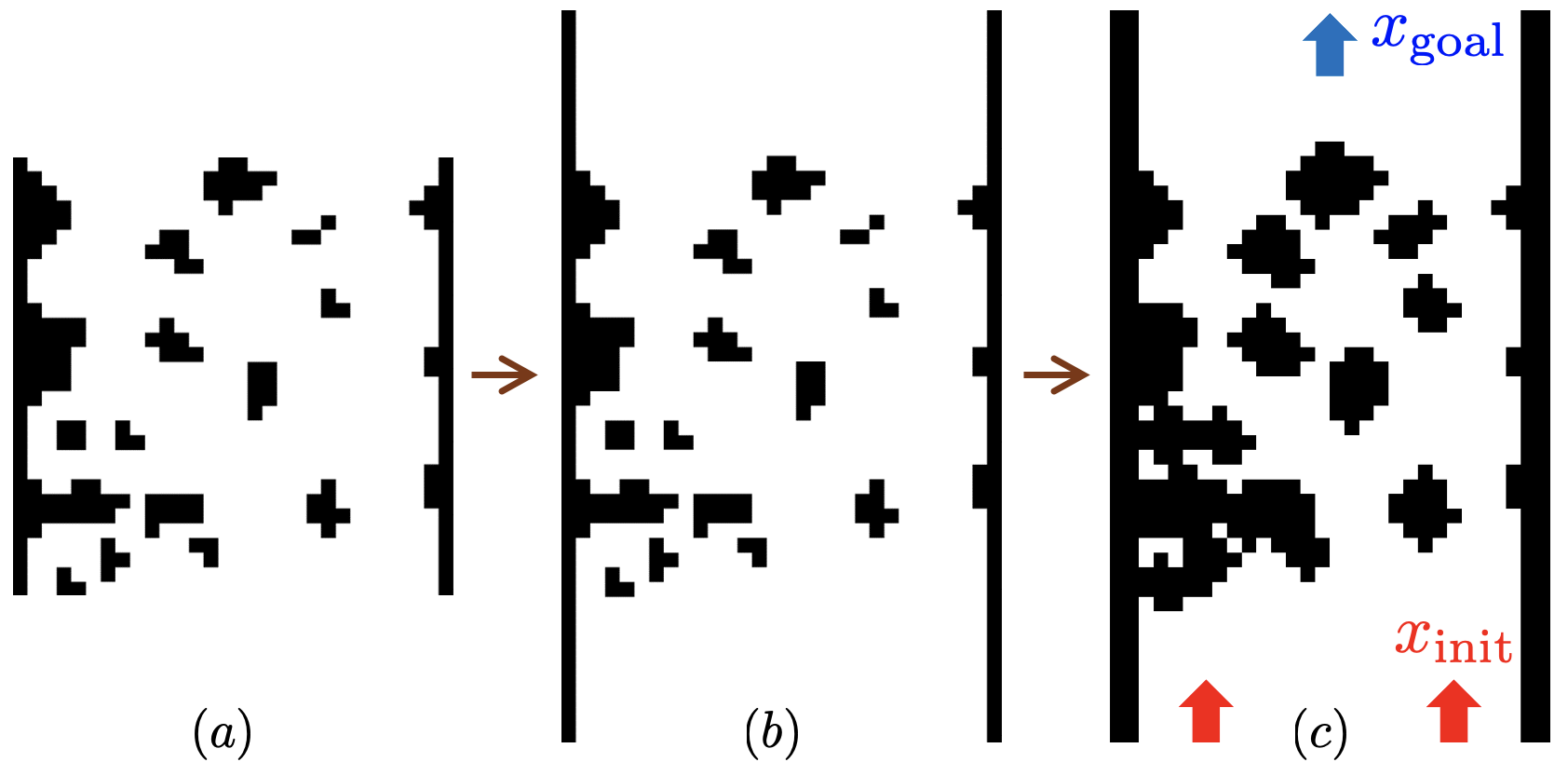}\vspace{-3mm}
    \caption{Map modification in the BARN dataset with extended boundaries and inflated obstacles: $(a)\rightarrow(b)\rightarrow(c)$. Red arrows mark the initial states, and the blue arrow marks the target state.}\vspace{-2.5mm}
    \label{fig:barn_modification}
\end{figure}

We use the BARN Dataset~\cite{perille2020benchmarking}, which contains 300 distinct 2D environments. The process of map modification is illustrated in Fig.~\ref{fig:barn_modification}. Each map is extended from 3m$\times$3m to 3m$\times$5m to create extra free space around the initial and target states, preventing collision.
The target state is set as \( x_{\rm goal} = [1.5, 5.0, \pi/2]^\top \), positioned in the upper middle of the map with an upward orientation. The two initial states, \( x_{\rm init} = [0.5, 0.0, \pi/2]^\top \) and \( x_{\rm init} = [2.5, 0.0, \pi/2]^\top \), are used for testing across the 300 different maps, leading to a total of 600 simulation trials. 

We compare our algorithm with several MPPI variants, including the original MPPI~\cite{williams2018information}, Log-MPPI~\cite{mohamed2022autonomous}, and Cluster-MPPI~\cite{patrick2024path}. All algorithms use the same control covariance \( \Sigma_u \) to ensure consistency. To account for the bidirectional nature of BiC-MPPI, we double the sample sizes \( N(N_{\rm f}) \) and extend the time horizon \( T(T_{\rm f}) \) for the other MPPI variants, providing a balanced comparison. For performance evaluation, a simulation is deemed successful if the Euclidean distance \( ||x_{\rm goal} - x_0|| < err \), with \( err = 0.1 \). If the robot fails to reach the goal within 200 iterations, which allows ample time for the robot to search the environment, the simulation is considered a failure.

%
All MPPI variants leverage warm-starting by using the previously optimized control input for forward rollouts in subsequent tasks. Table~\ref{tab:parameter} summarizes the key parameter settings for the different algorithms.
The results are presented in Table~\ref{tab:result}. While our method is not the fastest in terms of computation time, it exhibits superior reliability with the fewest failures and competitive computation times. Moreover, it consistently minimizes the number of iterations required to guide the robot to the goal.

\begin{table}[t]
	\caption{Parameters of MPPI algorithms in a wheeled mobile robot testing}\vspace{-4mm}
	\begin{center}
    \renewcommand{\arraystretch}{1.05}
    \begin{tabular}{|@{\,\,}c@{\,\,}|@{\,\,\,\,\,\,\,\,\,}c@{\,\,\,\,\,\,\,\,\,}|@{\,\,\,\,}c@{\,\,\,\,}|@{\,\,}c@{\,\,}|c|}
        \hline
        & MPPI & Log-MPPI & Cluster-MPPI & BiC-MPPI (ours)
        \\ \hline
        $\Sigma_u$ & \multicolumn{4}{c|}{diag([0.25, 0.25])} \\ \hline
        $\gamma_u$ & \multicolumn{4}{c|}{10} \\ \hline
        $T_{\rm f}$ & \multicolumn{3}{c|}{100} & 50 \\ \hline
        $T_{\rm b}$ & n/a & n/a & n/a& 50 \\ \hline
        $N_{\rm f}$ & \multicolumn{3}{c|}{6000} & 3000 \\ \hline
        $N_{\rm b}$ & n/a & n/a & n/a & 3000 \\ \hline
        $N_{\rm g}$ & n/a & n/a & n/a & 3000 \\ \hline
        $p$ & n/a & n/a & \multicolumn{2}{c|}{5} \\ \hline
        $\epsilon_{\max}$ & n/a & n/a & \multicolumn{2}{c|}{0.01} \\ \hline
        \multicolumn{5}{l}{\footnotesize \tiny n/a: not applicable} \hfill \\
    \end{tabular}
	\end{center}
	\vspace{-2mm}
	\label{tab:parameter}
\vspace{1mm}
	\caption{Results of simulation in BARN Dataset~\cite{perille2020benchmarking}}\vspace{-4mm}
	\begin{center}
    \renewcommand{\arraystretch}{1.05}
    \begin{tabular}{|@{\,}c@{\,}|@{\,\,\,\,\,\,}c@{\,\,\,\,\,\,}|@{\,\,}c@{\,\,}|@{\,\,}c@{\,\,}|@{\,\,}c@{\,\,}|}
        \hline
        & MPPI & Log-MPPI & Cluster-MPPI & BiC-MPPI (ours)
        \\ \hline
        No. Sim. & \multicolumn{4}{c|}{600} \\ \hline
        No. Failure & 117 & 107 & 119 & 79 \\ \hline
        Success Rate & 0.805 & 0.822 & 0.802 & 0.868 \\ \hline
        Avg. Iter. & 101.650 & 103.919 & 101.004 & 88.198 \\ \hline
        Avg. Time [s] & 2.766 & 4.161 & 7.356 & 3.346 \\ \hline
    \end{tabular}
	\end{center}
	\vspace{-2mm}
	\label{tab:result}
\vspace{3mm}
    \caption{Results of Quadrotor Landing}\vspace{-4mm}
    \begin{center}
    \renewcommand{\arraystretch}{1.05}
    \begin{tabular}{|@{\,}c@{\,}|@{\,\,\,\,\,\,}c@{\,\,\,\,\,\,}|@{\,\,\,\,}c@{\,\,\,\,}|@{\,\,}c@{\,\,}|@{\,\,}c@{\,\,}|}
        \hline
        & MPPI & Log-MPPI & Cluster-MPPI & BiC-MPPI (ours) \\ \hline
        No. Sim. & \multicolumn{4}{c|}{300} \\ \hline
        No. Failure & 78 & 170 & 81 & 10 \\ \hline
        Succ. Rate & 0.740 & 0.433 & 0.730 & 0.967 \\ \hline
        Avg. Iter. & 82.302 & 99.946 & 84.954 & 35.866 \\ \hline
        Avg. Time [s] & 5.642 & 10.437 & 7.768 & 2.456 \\ \hline
        Avg. Err & 2.105 & 6.592 & 2.011 & 0.531 \\ \hline
        Q1 & 0.773 & 4.500 & 0.829 & 0.263 \\ \hline
        Q2 & 1.367 & 6.604 & 1.359 & 0.433 \\ \hline
        Q3 & 2.694 & 8.821 & 2.490 & 0.692 \\ \hline
    \end{tabular}
    \end{center}
    \vspace{-1mm}
     {\footnotesize \tiny * Q1, Q2, Q3 represent the quartiles of terminal position error $| (x, y) - (x_{\rm \tiny{goal}}, y_{\rm \tiny{goal}})| $. \\[-1mm]
     ** Due to the fact that we didn't set constraint in the $y$-direction of map, errors could exceed 5.0.} 
    \label{tab:result_quadrotor}\vspace{-4mm}
\end{table}

\subsection{Quadrotor Landing in a Cluttered Space}
To evaluate the BiC-MPPI algorithm in a more complex scenario, we applied it to a quadrotor landing problem in a 3-dimensional environment. The continuous dynamics of the point-mass quadrotor are described by:
\begin{equation*}
	\begin{aligned}
		\dot{x} &= v_t, \
		\dot{v} &= a_t - g e_3,
	\end{aligned}
\end{equation*}
where the state vector \( [x_t^\top, v_t^\top]^\top \) consists of position \( x_t \in \mathbb{R}^3 \) and velocity \( v_t \in \mathbb{R}^3 \) across the \(x\), \(y\), and \(z\) axes. Here, \( g \) is the gravitational acceleration, \( e_3 \) is the unit vector along the \(z\)-axis (\( [0,0,1]^\top \)), and \( a_t \in \mathbb{R}^3 \) represents the control input in terms of acceleration.

The simulation uses a vertically extended version of the map from Fig.~\ref{fig:barn_modification}(c), scaling it to 3m$\times$5m$\times$5m. The initial state is set as \( x_{\rm init} = [1.5, 0, 5.0, 0, 0, 0]^\top \), with the goal of landing at \( x_{\rm goal} = [1.5, 5.0, 0, 0, 0, 0]^\top \). The experiment is considered a success if the quadrotor lands before reaching the iteration limit of 200 or colliding with an obstacle. A failure occurs if a collision is detected or if the maximum iteration limit is exceeded. The simulation setup and parameters mirror those of the previous 2D case, except where specified.

The system constraints for the quadrotor landing problem are defined as follows:
\begin{equation*}
\begin{split}
\mathcal{U} &= \{ a_t \in \mathbb{R}^3 : \| a_t \|_2 \leq a_{\rm max}, \| a_t \|_2 \cos(\theta_{\rm max}) \leq e_3^\top a_t \}, \\
\mathcal{X} &= \{ p \in \mathbb{R}^3 : p = (x_t, y_t, z_t) \notin \mathcal{O} \},
\end{split}
\end{equation*}
where the input constraints include a maximum acceleration \( a_{\rm max} = 20 \) and a thrust angle limit \( \theta_{\rm max} = 60^\circ \). The first condition restricts the acceleration norm, while the second limits the thrust angle relative to the \(z\)-axis. The state constraint \( \mathcal{X} \) ensures that the position \( (x_t, y_t, z_t) \) avoids obstacles \( \mathcal{O} \). As the map is vertically stacked, obstacle avoidance only requires checking \( x_t \) and \( y_t \) for collisions, similar to the 2D case.

The summary of the quadrotor landing results is shown in Table~\ref{tab:result_quadrotor}, where our BiC-MPPI consistently outperforms other MPPI variants across all performance metrics. The results highlight the algorithm's ability to handle increased complexity in both environmental and dynamic factors, demonstrating that as the problem becomes more challenging, BiC-MPPI becomes even more effective at optimizing trajectories and ensuring successful landings. This further validates the robustness of our method in complex 3D environments.

\section{Conclusions and Future Work}
\label{sec:conclusion}
This paper introduces the Bidirectional Clustered MPPI (BiC-MPPI) algorithm, a novel goal-oriented path integral optimizer. BiC-MPPI integrates kinematic and dynamic constraints, along with initial and terminal state conditions, into a framework capable of managing state and input constraints. Evaluations conducted on the BARN dataset, involving 600 simulations, revealed a $5.60\:\%$ to $8.23\:\%$ improvement in success rates compared to existing MPPI variants, while maintaining competitive computation times. In more complex 3D quadrotor landing simulations, BiC-MPPI achieved even more significant performance gains, outperforming three MPPI variants with success rates increasing by $30.68\:\%$ to $123.33\:\%$, excelling in both collision avoidance and computational efficiency. Future work will explore adapting BiC-MPPI for real-time feedback, dynamic environments, moving targets, and scenarios involving multiple intermediate waypoints.

\phantom{
\cite{patrick2024path}
\cite{nayak2022bidirectional}
\cite{ester1996density}
\cite{kazim2024recent}
\cite{tedrake2009lqr}
\cite{kim2024mppi}
\cite{trevisan2024biased}
\cite{butcher2016numerical}
\cite{bian2018survey}
\cite{perille2020benchmarking}
\cite{williams2018information}
\cite{mohamed2022autonomous}
\cite{chai2020overview}
\cite{ART002429316}
}

\newpage

\bibliographystyle{IEEEtran}
\bibliography{bi_mppi}

\begin{thebibliography}{10}
\providecommand{\url}[1]{#1}
\csname url@samestyle\endcsname
\providecommand{\newblock}{\relax}
\providecommand{\bibinfo}[2]{#2}
\providecommand{\BIBentrySTDinterwordspacing}{\spaceskip=0pt\relax}
\providecommand{\BIBentryALTinterwordstretchfactor}{4}
\providecommand{\BIBentryALTinterwordspacing}{\spaceskip=\fontdimen2\font plus
\BIBentryALTinterwordstretchfactor\fontdimen3\font minus
  \fontdimen4\font\relax}
\providecommand{\BIBforeignlanguage}[2]{{%
\expandafter\ifx\csname l@#1\endcsname\relax
\typeout{** WARNING: IEEEtran.bst: No hyphenation pattern has been}%
\typeout{** loaded for the language `#1'. Using the pattern for}%
\typeout{** the default language instead.}%
\else
\language=\csname l@#1\endcsname
\fi
#2}}
\providecommand{\BIBdecl}{\relax}
\BIBdecl

\bibitem{choset2005principles}
H.~Choset, K.~M. Lynch, S.~Hutchinson, G.~A. Kantor, and W.~Burgard,
  \emph{Principles of Robot Motion: Theory, Algorithms, and
  Implementations}.\hskip 1em plus 0.5em minus 0.4em\relax MIT Press, 2005.

\bibitem{lavalle2006planning}
S.~M. LaValle, \emph{Planning Algorithms}.\hskip 1em plus 0.5em minus
  0.4em\relax Cambridge University Press, 2006.

\bibitem{chai2020overview}
R.~Chai, A.~Savvaris, A.~Tsourdos, and S.~Chai, ``Overview of trajectory
  optimization techniques,'' in \emph{Design of Trajectory Optimization
  Approach for Space Maneuver Vehicle Skip Entry Problems}.\hskip 1em plus
  0.5em minus 0.4em\relax Springer, 2020, pp. 7--25.

\bibitem{kazim2024recent}
M.~Kazim, J.~Hong, M.-G. Kim, and K.-K.~K. Kim, ``Recent advances in path
  integral control for trajectory optimization: {A}n overview in theoretical
  and algorithmic perspectives,'' \emph{Annual Reviews in Control}, vol.~57, p.
  100931, 2024.

\bibitem{jordan2013optimal}
M.~Jordan and A.~Perez, ``Optimal bidirectional rapidly-exploring random
  trees,'' MIT, Tech. Rep., 2013.

\bibitem{osborne1969shooting}
M.~R. Osborne, ``On shooting methods for boundary value problems,''
  \emph{Journal of Mathematical Analysis and Applications}, vol.~27, no.~2, pp.
  417--433, 1969.

\bibitem{keller2018numerical}
H.~B. Keller, \emph{Numerical methods for two-point boundary-value
  problems}.\hskip 1em plus 0.5em minus 0.4em\relax Courier Dover Publications,
  2018.

\bibitem{nayak2022bidirectional}
S.~Nayak and M.~W. Otte, ``Bidirectional sampling-based motion planning without
  two-point boundary value solution,'' \emph{IEEE Transactions on Robotics},
  vol.~38, no.~6, pp. 3636--3654, 2022.

\bibitem{kim2024mppi}
M.-G. Kim, M.~Jung, J.~Hong, and K.-K.~K. Kim, ``{MPPI-IPDDP}: {H}ybrid method
  of collision-free smooth trajectory generation for autonomous robots,''
  \emph{arXiv preprint arXiv:2208.02439}, 2024.

\bibitem{trevisan2024biased}
E.~Trevisan and J.~Alonso-Mora, ``{Biased-MPPI}: {I}nforming sampling-based
  model predictive control by fusing ancillary controllers,'' \emph{IEEE
  Robotics and Automation Letters}, 2024.

\bibitem{bauschke1996projection}
H.~H. Bauschke and J.~M. Borwein, ``On projection algorithms for solving convex
  feasibility problems,'' \emph{SIAM review}, vol.~38, no.~3, pp. 367--426,
  1996.

\bibitem{patrick2024path}
S.~Patrick and E.~Bakolas, ``Path integral control with rollout clustering and
  dynamic obstacles,'' \emph{arXiv preprint arXiv:2403.18066}, 2024.

\bibitem{ester1996density}
M.~Ester, H.-P. Kriegel, J.~Sander, X.~Xu \emph{et~al.}, ``A density-based
  algorithm for discovering clusters in large spatial databases with noise,''
  in \emph{Knowledge Discovery and Data Mining (KDD)}, vol.~96, no.~34, 1996,
  pp. 226--231.

\bibitem{butcher2016numerical}
J.~C. Butcher, \emph{Numerical methods for ordinary differential
  equations}.\hskip 1em plus 0.5em minus 0.4em\relax John Wiley \& Sons, 2016.

\bibitem{tedrake2009lqr}
R.~Tedrake, ``{LQR-trees}: {F}eedback motion planning on sparse randomized
  trees.'' in \emph{Robotics: Science and Systems}, vol. 2009, 2009.

\bibitem{perille2020benchmarking}
D.~Perille, A.~Truong, X.~Xiao, and P.~Stone, ``Benchmarking metric ground
  navigation,'' in \emph{IEEE International Symposium on Safety, Security, and
  Rescue Robotics (SSRR)}.\hskip 1em plus 0.5em minus 0.4em\relax IEEE, 2020,
  pp. 116--121.

\bibitem{williams2018information}
G.~Williams, P.~Drews, B.~Goldfain, J.~M. Rehg, and E.~A. Theodorou,
  ``Information-theoretic model predictive control: {T}heory and applications
  to autonomous driving,'' \emph{IEEE Transactions on Robotics}, vol.~34,
  no.~6, pp. 1603--1622, 2018.

\bibitem{mohamed2022autonomous}
I.~S. Mohamed, K.~Yin, and L.~Liu, ``Autonomous navigation of {AGV}s in unknown
  cluttered environments: {L}og-{MPPI} control strategy,'' \emph{IEEE Robotics
  and Automation Letters}, vol.~7, no.~4, pp. 10\,240--10\,247, 2022.

\bibitem{bian2018survey}
J.~Bian, D.~Tian, Y.~Tang, and D.~Tao, ``A survey on trajectory clustering
  analysis,'' \emph{arXiv preprint arXiv:1802.06971}, 2018.

\bibitem{ART002429316}
H.-H. Kwon, H.-S. Shin, Y.-H. Kim, and D.-H. Lee, ``Trajectory optimization for
  impact angle control based on sequential convex programming,'' \emph{The
  Transactions of The Korean Institute of Electrical Engineers (in Korean)},
  vol.~68, no.~1, pp. 159--166, 2019.

\end{thebibliography}

\balance

\end{document}